\documentclass{article}
\pdfoutput=1
\usepackage{arxiv}
\usepackage[utf8]{inputenc} 
\usepackage[T1]{fontenc}    
\usepackage{hyperref}       
\usepackage{url}            
\usepackage{booktabs}       
\usepackage{amsfonts}       
\usepackage{nicefrac}       
\usepackage{microtype}      
\usepackage{lipsum}
\usepackage{fancyhdr}       
\usepackage{graphicx}       
\usepackage{subcaption}
\usepackage{natbib}
\setcitestyle{numbers,square}
\graphicspath{{media/}}     
\usepackage{multirow}
\usepackage{algorithm}
\usepackage{algorithmic} 
\usepackage[utf8]{inputenc} 
\usepackage[T1]{fontenc}    
\usepackage{amsmath}        
\usepackage{stmaryrd}
\usepackage{xcolor}
\definecolor{red}{HTML}{ef5734}
\definecolor{green}{HTML}{006837}
\definecolor{blue}{HTML}{075aaa}

\title{Maintaining Informative Coherence: Migrating Hallucinations in Large Language Models via Absorbing Markov Chains
}

\author{
  Jiemin Wu\\
  HKUST(GZ)\\
  Guangzhou, China\\
  \texttt{jieminwu@hkust-gz.edu.cn}\\
  \And
  Songning Lai\\
  HKUST(GZ)\\
  Guangzhou, China\\
  \texttt{songninglai@hkust-gz.edu.cn}\\
  \And
  Ruiqiang Xiao\\
  HKUST(GZ)\\
  Guangzhou, China\\
  \texttt{ruiqiangxiao@hkust-gz.edu.cn}\\
  \And
  Tianlang Xue\\
  HKUST(GZ)\\
  Guangzhou, China\\
  \texttt{tianlangxue@hkust-gz.edu.cn}\\
  \And
  Jiayu Yang\\
  HKUST(GZ)\\
  Guangzhou, China\\
  \texttt{jyang729@connect.hkust-gz.edu.cn}\\
  \And
  Yutao Yue\thanks{Correspondence to Yutao Yue \{yutaoyue@hkust-gz.edu.cn\}}\\
  HKUST(GZ)\\
  Guangzhou, China\\
  \texttt{yutaoyue@hkust-gz.edu.cn}\\
  \thanks{This work was supported by Guangzhou-HKUST(GZ) Joint Funding Program(Grant No.2023A03J0008), Education Bureau of Guangzhou Municipality}
}

\begin{document}
\maketitle

\begin{abstract}
Large Language Models (LLMs) are powerful tools for text generation, translation, and summarization, but they often suffer from hallucinations-instances where they fail to maintain the fidelity and coherence of contextual information during decoding, sometimes overlooking critical details due to their sampling strategies and inherent biases from training data and fine-tuning discrepancies. These hallucinations can propagate through the web, affecting the trustworthiness of information disseminated online. To address this issue, we propose a novel decoding strategy that leverages absorbing Markov chains to quantify the significance of contextual information and measure the extent of information loss during generation. By considering all possible paths from the first to the last token, our approach enhances the reliability of model outputs without requiring additional training or external data. Evaluations on datasets including TruthfulQA, FACTOR, and HaluEval highlight the superior performance of our method in mitigating hallucinations, underscoring the necessity of ensuring accurate information flow in web-based applications.
\end{abstract}


\section{Introduction}
With the advent of Large Language Models (LLMs), a transformative era in natural language processing (NLP) has emerged, marking significant advancements in the capabilities of text generation \citep{li2024dtllm}, translation \citep{zhang2023prompting}, summarization \citep{ravaut2024context}, and more \citep{min2023recent, peng2023study, demszky2023using}. Built upon deep learning architectures such as transformers \citep{vaswani2017attention}, these models have demonstrated exceptional proficiency in producing text that mirrors human-like fluency and complexity \citep{chen2024large, li2024dtllm}. The rise of LLMs has profoundly impacted the internet ecosystem, enhancing the ability to generate, translate, and summarize content at scale, thereby reshaping how information is created and shared online.

\begin{figure*} [tbp]
    \centering
    \includegraphics[width=0.8\linewidth]{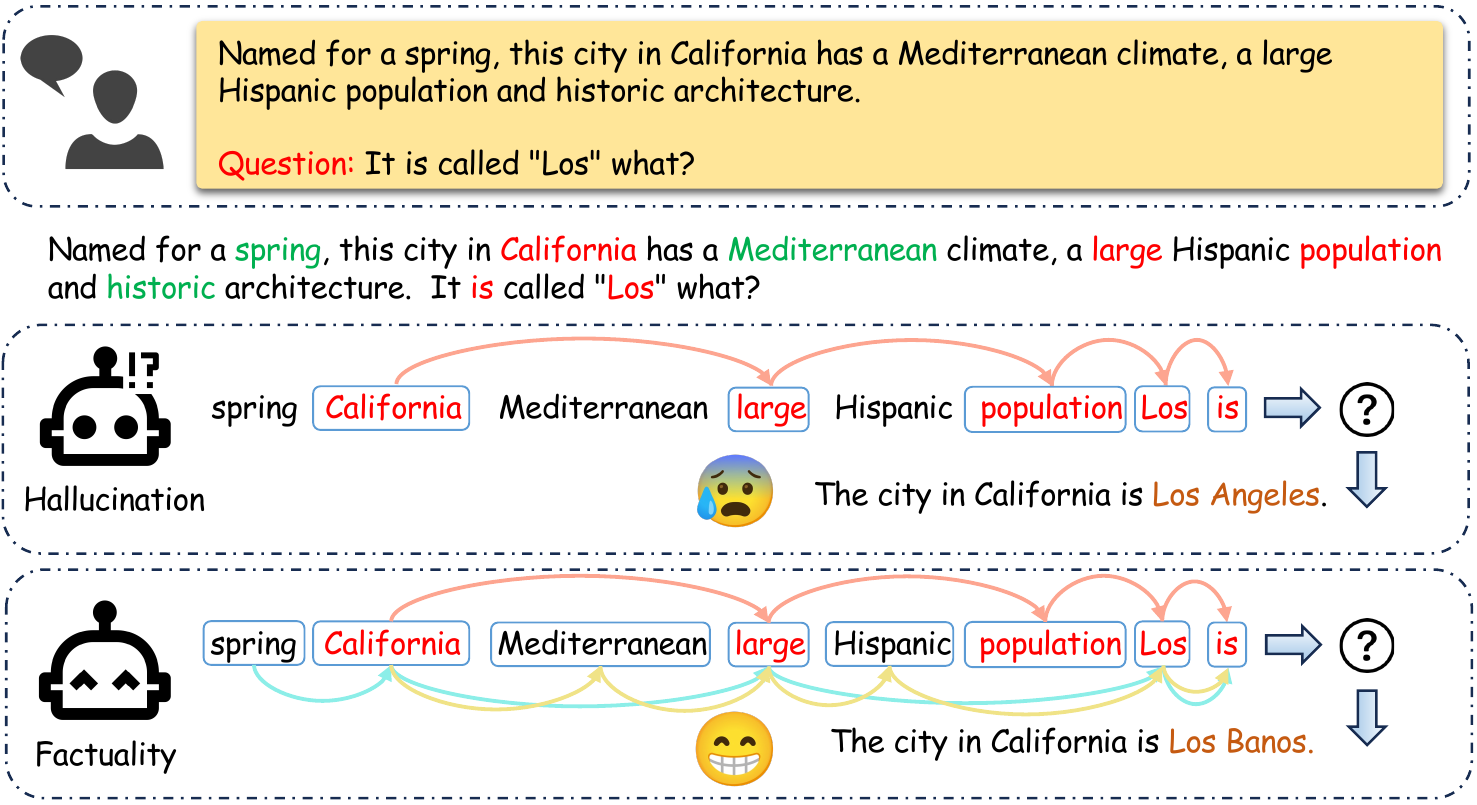}
    \caption{Hallucinations phenomenon in large models due to overlooking key context. The top section shows a complete prompt and question. Important tokens are highlighted in red, and less critical ones in green. Hallucinations potentially arise from reliance on red tokens while ignoring green tokens.}
    \label{fig:introduction}
\end{figure*}

However, alongside these advancements, LLMs have introduced a notable challenge: the generation of hallucinations—outputs that, while linguistically coherent, are ungrounded or factually inaccurate \citep{Zhang2023SirensSI}. Such inaccuracies can propagate through the web, eroding the trustworthiness of information and posing significant risks in domains where precision and authenticity are paramount \citep{tonmoy2024comprehensive}. For instance, consider the simple question, "What is the capital of Australia?" the correct answer is "Canberra," but an LLM might generate a coherent yet incorrect answer, such as "Sydney," due to its reliance on statistical patterns and potential biases in the training data. This propensity not only questions the reliability of LLMs but also highlights the need for robust mechanisms to ensure the accuracy and coherence of generated text.

The generation of hallucinations in LLMs is a multifaceted issue, rooted in the models' reliance on statistical patterns within training data, lacking explicit understanding of the content or context \citep{pan2024not}. While LLMs are generally adept at distinguishing between reliable and unreliable information, they sometimes struggle to do so effectively, especially when influenced by biases in training and fine-tuning data, the proximity of nearby words, and the randomness introduced by sampling strategies. These factors can cause the model to overlook critical information from earlier parts of the context when generating the next word.

To address this challenge, we propose a novel framework that leverages the theory of absorbing Markov chains \citep{grinstead2012introduction} to mitigate hallucinations. Our approach views the process of LLMs generating new content given a context as the propagation of information from the given context to the new content. Traditional autoregressive generation models each token's probability based on its entire preceding context. However, a more effective approach would consider the possibility of direct jumps from certain prefix tokens to specific keywords, similar to how observing "Australia" can immediately evoke "Canberra." By considering all possible information flow paths from the first to the last token, rather than just the immediate context, we can more comprehensively quantify the importance of key information and its flow.

We observe that the set of all paths from the source to the sink in this information flow can be modeled as an absorbing Markov chain, leveraging well-established theoretical foundations to precisely model key information and its flow. Specifically, we define the criticality of a token using a combination of absorbing Markov chains and information theory: if a token represents a state that is difficult to reach from its preceding states, it is deemed highly critical. This indicates that the token cannot be predicted from the prior context and introduces more new information to the context. For the flow of information, we calculate the probability of transitioning from the state corresponding to a preceding token to the state of the target token via random walks, characterizing the transfer of information from the preceding token to the target token.

During the generation of the next token, we compute the information value contained in each token within the context and the information loss after it flows to the last token. This loss reflects the degree to which key information is diminished as it reaches the final token. Consequently, we adjust the probability distribution of the next token based on the information loss from the preceding context and the initial token distribution, guiding the model to refocus on important contextual information that might have been overlooked, thereby alleviating hallucinations.

A key advantage of our approach, compared to other methods addressing hallucinations in large models, is its independence from accessing the internal representations of the model. This attribute ensures a more stable application across a variety of large model architectures without the need for parameter adjustments. Consequently, our method stands out as a scalable and model-agnostic solution, offering a significant contribution to the ongoing efforts to enhance the reliability and accuracy of LLM-generated content.

Our contributions are distinct and impactful, summarized as follows:
\begin{itemize}
\item We bridge the gap between LLM inference processes and absorbing Markov chains, offering a novel theoretical perspective for understanding and improving text generation.
\item Our method quantifies information loss during text generation, providing a metric to gauge and enhance coherence and reliability.
\item We introduce a mechanism to adjust token probabilities based on information loss, directly mitigating the generation of hallucinations and promoting the production of contextually relevant text.
\end{itemize}
By integrating this feedback mechanism into the token generation process, our methodology aims to steer LLMs toward generating outputs that better preserve the fidelity of the intended information flow, significantly reducing the risk of hallucinations and ensuring the integrity and reliability of the information landscape.

\section{Related Work}
\subsection{Hallucinations in LLMs}
Hallucinations in LLMs can arise from several sources, including massive training data that may introduce fabricated, outdated, or biased information \citep{Zhang2023SirensSI}. The versatility of LLMs, designed to excel in cross-task, cross-lingual, and cross-domain settings, complicates the comprehensive evaluation and mitigation of hallucinations \citep{tonmoy2024comprehensive}. Various techniques have been proposed to mitigate hallucinations, such as Retrieval-Augmented Generation (RAG) \citep{lewis2020retrieval}, Inference-time intervention \citep{li2024inference}, Knowledge Retrieval \citep{varshney2023stitch}, Self-Reflection \citep{ji2023towards}, Exploiting Uncertainty \citep{lin2023generating}, Chain-of-Thought Prompting \citep{wei2022chain}, and System Prompts \citep{touvron2023llama}. These methods aim to enhance the model's ability to ground its outputs in factual information and maintain consistency with the input context. Despite these advancements, the challenge of hallucinations in LLMs remains an active area of research, requiring further innovation and evaluation \citep{Zhang2023SirensSI, tonmoy2024comprehensive}.

\subsection{Constrained Decoding Strategies}
Many studies focus on using modified decoding strategies during inference to mitigate hallucinations, as adjusting model parameters requires extensive computational resources. Notable techniques include Context-Aware Decoding (CAD), which uses a contrastive output distribution to amplify the difference between the output probabilities with and without context, effectively overriding the model’s prior knowledge \citep{shi2023trusting}. Inference-Time Intervention (ITI) shifts model activations during inference to enhance truthfulness by identifying and shifting activations along attention heads with high linear probing accuracy for truthfulness \citep{li2024inference}. Decoding by Contrasting Layers (DOLA) contrasts logit differences between later and earlier layers to minimize the generation of incorrect facts \citep{chuang2023dola}. Additionally, Activation Decoding manipulates the activation patterns of the model during inference to guide the generation process toward more factual and contextually appropriate outputs by measuring and optimizing the sharpness of in-context activations \citep{chen2024context}. These strategies aim to improve the factuality and coherence of LLM outputs by guiding the generation process toward more reliable and contextually consistent results.

\subsection{Information Flow in LLMs}
Many studies have introduced the perspective of information flow to better understand the internal workings of large language models (LLMs). For instance, \citep{abnar2020quantifying} proposed methods to quantify attention flow in Transformers, showing that raw attention weights can be unreliable for explaining model decisions. They introduced attention rollout and attention flow as post-hoc methods to approximate the relevance of input tokens, yielding higher correlations with importance scores obtained using ablation methods and input gradients. Similarly, \citep{ferrando2024information} developed a method to automatically build graphs representing information flow routes within LLMs, tracing information back through the network and retaining only the most important nodes and edges to efficiently uncover existing circuits with a single forward pass. \citep{wang2023label} explored the role of label words as anchors in in-context learning (ICL), revealing that label words aggregate information from demonstration examples in shallow layers and serve as references for final predictions in deep layers. \citep{yao2024knowledge} delved into the computation graph of LLMs to uncover knowledge circuits that are instrumental in encoding specific knowledge, revealing how certain information heads, relation heads, and Multilayer Perceptrons (MLPs) collaboratively encode knowledge within the model. \citep{yuan2021explaining} proposed Transition Attention Maps to generate saliency maps concerning specific target categories in Vision Transformers, combining Markov chains and integrated gradients to investigate information flow across layers and compute the relevance of input tokens for model decisions. These studies collectively highlight the importance of understanding information flow in LLMs to enhance their interpretability and reliability.

\subsection{Absorbing Markov Chain }
AMCs are a special kind of Markov chain where some states are absorbing states that, once reached, cannot be left. Their mathematical foundation and applications have been extensively studied, providing valuable insights into various fields ranging from mathematics to engineering \cite{Kemeny1976}. In particular, AMCs have been applied to model and analyze systems with terminal states, which has implications for reliability and queueing theory \cite{Doyle1984, Stewart2009}. Despite these advancements, the application of absorbing Markov chains (AMCs) to understand the flow of information during inference in large language models (LLMs) remains underexplored. Hallucinations in LLMs occur when the model generates nonsensical or unrelated content, often due to the lack of strong contextual cues or the influence of biased training data. By conceptualizing the last token as an absorbing state and considering all possible paths from prefix contexts to subsequent tokens, our approach can quantify the informational content of contextual tokens. This enables the model to focus more on overlooked tokens during decoding, ensuring the generation of factually accurate and coherent content.

\section{Method}

Our approach aims to optimize the inference process of large language models by treating it as an information flow from the beginning to the end of the context. Specifically, we view the entire inference process as information traveling along various paths from the first token to the last token in the context. To more effectively capture and utilize these information flows, we propose a framework based on an absorbing Markov chain model.

\subsection{Language Model Architecture}
An autoregressive language model operates by predicting the probability distribution over the vocabulary for the next token given the history of previously generated tokens. Mathematically, this can be expressed as:

\begin{equation}
   \mathbb{P}(x_{t+1} | x_1, x_2, \ldots, x_t) 
   \label{eq1}
\end{equation}

\noindent where $ x_1, \ldots, x_t $ represent the input tokens up to time step $ t $, and $ x_{t+1} $ is the next token to be predicted. The model learns to map the sequence of input tokens to the probability distribution over the vocabulary, which is typically achieved using deep neural networks such as Transformers. Although in theory, the probability of each token can be determined by all previous tokens and the model, training biases can lead the model to overlook some important tokens in the context, thereby causing hallucinations. Therefore, a better approach is to consider the paths from all possible context prefixes to a given token, fully utilizing the prediction information generated during the inference process.

\subsection{Absorbing Markov Chain Formulation}
In our approach, we conceptualize the process of predicting tokens through random walks at each time step as being equivalent to the states in a Markov chain. We first introduce our absorbing Markov chain. Our AMC is denoted by $\mathbf{X}$, with the set of random variables $\mathbf{X} = \{X_n : n>0\}$. Within the probability space $\{\Omega, \mathcal{F}, \mathbb{P}\}$, it uses a one-dimensional countable set as the index set, and this state space is denoted by $\mathbf{s}$. The state space consists of the various states $s_i$, which are the predicted tokens in past $t$ time steps. In our ACM, the random variables satisfies:
\begin{equation}
    \mathbb{P}\{X_{t+1} | X_t, \ldots, X_1\} = \mathbb{P}\{X_{t+1} | X_t\}.
\end{equation}
In the state space $\mathbf{s}$ of our AMC, each state represents the token that have appeared before time step $t+1$.

To model the inference process more effectively, we use an AMC represented by a modified transition probability matrix $\tilde{P}$ which expanded upon original transition probability matrix $P$. Let $P$ be the transition probability matrix of a AMC in autoregressive language model, defined as follows:
The matrix $P$ is an $n \times n$ square matrix, where each element $p_{ij}$ represents the conditional probability of transitioning from state $i$ to state $j$, that is:
$$
p_{ij} = \mathbb{P}(X_{t+1} = j \mid X_t = i)
$$
Here, $X_t$ denotes the state of the Markov chain at time $t$, which mapping to the predicted token at time step $t$. When $p_{i\cdot} = \sum_{j=1}^{t}p_{ij} = 0$, this implies that when the Markov chain enters state $i$, it cannot transition to any other state. We refer to this type of state as an absorbing state.
This matrix captures the transitions between tokens and incorporates mechanisms to account for information loss:

\textbf{1) Matrix Construction}: We modify the original transition probability matrix $P$ to $\tilde{P}$ by setting $\tilde{P}_{tt} = 1$ to introduce absorbing states, ensuring that the sum of each row equals 1. This guarantees that once the chain reaches an absorbing state, it remains there.

\textbf{2) Submatrices Definition}: We partition $\tilde{P}$ into submatrices to differentiate between transient and absorbing states:

\begin{equation}
    \tilde{P} = \begin{bmatrix} Q & R \\ \mathbf{0} & I_{1} \end{bmatrix},
    \label{eq2}
\end{equation}

\noindent where $Q$ is a $t \times t$ matrix representing transitions among transient states (tokens), $R$ is a $t \times 1$ matrix representing transitions from transient states to the absorbing state, $\mathbf{0}$ is a $1 \times t$ zero matrix and $I_{1}$ is a $1 \times 1$ identity matrix for the absorbing state. 

\textbf{3) Fundamental Matrix $N$}: The fundamental matrix $N$ is computed as:

\begin{equation}
    N = (I_t - Q)^{-1}
    \label{eq3}
\end{equation}

\noindent where $I_t$ represents a t-dimensional identity matrix corresponding to the number of transient states. This matrix $N$ gives the expected number of times the chain will be in each transient state before absorption.

\textbf{4) Visitation Probabilities $V$}: In the context of absorbing Markov chains, the matrix $V$ represents the visitation probabilities of transient states before reaching an absorbing state. Specifically, when starting from a transient state $i$, the probability of visiting another transient state $j$ is given by the $(i,j)$-entry of matrix $V$. This matrix is crucial for understanding the flow of information through the states (or tokens, in the case of LLMs) and for identifying where information loss is more pronounced.

After deriving the fundamental matrix $N$, the matrix $V$ is then obtained by normalizing $N$ to represent probabilities rather than expected counts. Formally, $V$ is computed as follows:

\begin{equation}
    V = N \cdot (D_N)^{-1},
    \label{eq4}
\end{equation}

\noindent where $D_N$ is a diagonal matrix with the same diagonal elements as $N$. $V(t)$, specifically, refers to the $t$-th column of the matrix $V$, representing the probabilities of visiting all transient states from the transient state corresponding to the $t$-th token before reaching an absorbing state.

Now, we will present the properties of matrix $P$ and its expansion matrix $\tilde{P}$, and demonstrate the validity of their construction.

\textbf{i. Upper triangular matrix $P$:} In our transition probability matrix $P$, $p_{ij} = 0, i > j$. 
Unlike the conventional Markov transition probability matrix, in ours, the probabilities are zero when $i > j$. This is due to the masking mechanism in the transformer, where only the preceding tokens influence the prediction of the next token.

\textbf{ii. Interpretations of diagonal elements:} Distinct from conventional Markov transition matrices, it should be emphasized that the values of the diagonal elements in our matrix do not signify the probability of a state returning to itself. Consequently, our Markov chain has no recurrent states. This is because LLMs introduce unique positional encodings for tokens at different positions, ensuring that even identical tokens have distinct semantics at different locations, thus precluding the possibility of returning to the same state. However, to ensure the theoretical integrity of the Markov chain, we uphold the completeness of the following properties:
$$
\sum_{j=1}^{t} p_{ij} = 1, i=1, 2, \ldots, t.
$$
Here, we ensure that the sum of transition probabilities for each row equals one. Therefore, the values on the diagonal can be interpreted as the probability of transitioning from the current state to tokens outside the subsequent context.

\begin{figure*} [htbp]
    \centering
    \includegraphics[width=0.8\linewidth]{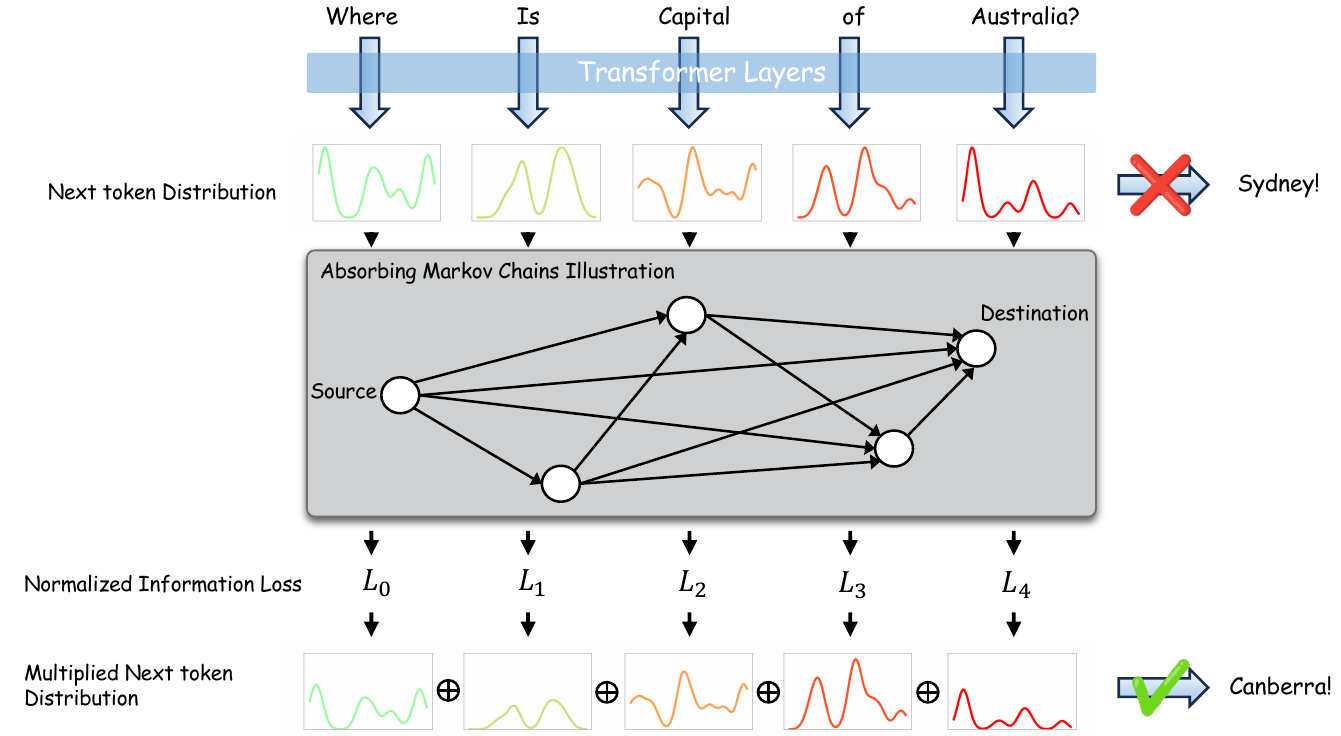}
    \caption{Illustration of the conceptual diagram for a token-based absorbing Markov chain. The figure depicts the interconnections among four transient states and one absorbing state.}
    \label{fig:Method}
\end{figure*}

\subsection{Quantification of Information Flow}

The core of our approach lies in quantifying token-$i$'s flow and loss of information through the computation of the information score $S(i)$ and information loss $\mathcal{L}_{\text{info}}(i)$. These metrics help us understand how well the model retains and propagates key information during the generation process.

\textbf{1) Information Score $S(i)$ of the $i$-th token}: Define the information score as
\begin{equation}
    S(i) = - \log V_{1, i}
    \label{eq5}
\end{equation}

\noindent where $V_{1, i}$ represents the probabilities of visiting the transient states of the $i$-th token from the initial transient state before reaching an absorbing state. The information score $S(i)$ provides a measure of the criticality of the current state. A high value of $S(i)$ indicates that the current token is difficult to reach from its preceding states and contains more information.

\textbf{2) Information Loss $\mathcal{L}_{\text{info}}(i)$}: Define the information loss as
\begin{equation}
    \mathcal{L}_{\text{info}}(i) = S(i) \cdot (1 -  V_{i, t})
    \label{eq6}
\end{equation}

\noindent where $V$ represents the transient visitation probabilities. The information loss $\mathcal{L}_{\text{info}}(i)$ quantifies the degree to which the information from previous states is lost when transitioning to the current state. Specifically, it directly measures the transfer of information scores from all contextual tokens to the last token, reflecting the extent of information loss up to the last token. A high value of $\mathcal{L}_{\text{info}}(i)$ indicates significant information loss, suggesting that the semantic up to the last token does not effectively preserve the key information from the previous context.

By using these metrics, we can identify tokens that are critical for maintaining the context and detecting where information loss occurs. This helps in mitigating hallucinations by ensuring that the model generates content that is more aligned with the key information in the input context.

\subsection{Adjusting Token Probabilities}
To mitigate hallucinations, we adjust the token probability distributions based on the information loss $\mathcal{L}_{\text{info}}(t)$. The adjusted distribution $\tilde{D}(t)$ for the $t$-th token is computed as:
\begin{equation}
    \tilde{D}(t) = D(t) + \lambda \left( \sum_{i=0}^{t-1} \mathbf{Norm}(\mathcal{L}_{\text{info}}(i) \cdot D(i) )\right),
    \label{eq7}
\end{equation}

\noindent where $\lambda$ is a tuning parameter that controls the adjustment extent.

Overall, our framework offers a structured approach to understanding and reducing hallucinations in LLMs by leveraging the theory of absorbing Markov chains to model the inference process. By quantifying information flow and loss, and adjusting token distributions accordingly, we aim to enhance the model's fidelity and coherence. The pseudocode is shown in Algorithm \ref{slg:pse}.

\begin{algorithm}
\caption{Adjusting Token Probabilities using Absorbing Markov Chain}
\label{slg:pse}
\begin{algorithmic}
\REQUIRE $x_1, x_2, \ldots, x_t$ $\gets$ input tokens
\REQUIRE $P$ $\gets$ transition probability matrix
\REQUIRE $\lambda$ $\gets$ tuning parameter
\REQUIRE $D(t)$ $\gets$ original token probability distribution at time step $t$
\ENSURE $\tilde{D}(t)$ $\gets$ adjusted token probability distribution at time step $t$

\STATE // Construct the original transition probability matrix $P$ to $\tilde{P}$
\STATE $\tilde{P} = \begin{bmatrix} Q & R \\ \mathbf{0} & I_{1} \end{bmatrix}$

\STATE // Compute the fundamental matrix $N$
\STATE $N = (I_t - Q)^{-1}$

\STATE // Compute the visitation probabilities $V$
\STATE $V = N \cdot (D_N)^{-1}$

\STATE // Initialize the adjusted token probability distribution $\tilde{D}(t)$
\STATE $\tilde{D}(t) = D(t)$

\STATE // Iterate over each time step $t$
\FOR{$i = 1$ to $t$}
  \STATE // Compute the information score $S(i)$
  \STATE $S(i) = - \log V_{1, i}$

  \STATE // Compute the information loss $\mathcal{L}_{\text{info}}(t)$
  \STATE $\mathcal{L}_{\text{info}}(i) = S(i) \cdot (1 -  V_{i, t})$

\ENDFOR

  \STATE // Adjust the token probability distribution $\tilde{D}(t)$
  \STATE $\tilde{D}(t) = D(t) + \lambda \left( \sum_{i=0}^{t-1} \mathbf{Norm}(\mathcal{L}_{\text{info}}(i) \cdot D(i) )\right)$

\RETURN $\tilde{D}(t)$
\end{algorithmic}
\end{algorithm}

\subsection{Further elaboration of the methodology analysis}

\noindent\textbf{Why do we use the Markovian framework?} By structuring token transitions as a Markov process, we inherently prioritize coherence and contextuality in text generation considering all possible information flow.  The Markovian framework naturally lends itself to capturing dependencies and sequential relationships, which are critical for producing coherent text. The methodology introduces a quantitative measure of information flow and loss, enabling a more rigorous approach to diagnosing and mitigating hallucinations.  This quantification is a step towards making the often qualitative assessments of text generation quality more objective and actionable.

\noindent\textbf{Why this method could quantify information loss?} The fundamental matrix $\mathbf{N}$, derived from the submatrix $Q$ of $\tilde{P}$, encapsulates the expected number of times the process visits each transient state before reaching an absorbing state. This matrix is pivotal for understanding how information propagates through the model. The information score $S(t)$ and the information loss $\mathcal{L}_{\text{info}}(t)$ are derived from $\mathbf{N}$, providing a measure of how much information is retained or lost as tokens are generated.

\noindent\textbf{Why are we adjusting token probabilities?} The primary rationale for adjusting token probabilities based on information loss is to minimize the divergence from the expected information flow, thereby reducing the likelihood of hallucinatory content. By preferentially selecting tokens that contribute to a lower information loss, the model is guided towards more coherent and contextually relevant generations. 

The adjustment process involves adding a weighted sum of the predicted next token distributions given all possible prefix context to the original distribution $D(t)$ according to their information loss. This operation not only penalizes tokens associated with higher information loss but also normalizes the distribution to maintain probabilistic integrity. The parameter $\lambda$ allows for the tuning of this adjustment, striking a balance between adherence to the original model predictions and the corrective influence of the information loss metric.

From a theoretical perspective, this approach introduces a feedback mechanism into the token generation process, where the selection of each token is informed by both its immediate probability and its contribution to the overall coherence of the text as quantified by the Markov chain model. This feedback loop is a manifestation of introducing external, quantifiable knowledge (the structure of absorbing Markov chains and information theory) into the inherently probabilistic process of language generation.

\section{Experiments}

\label{sec:exp}
\subsection{Setup}
\noindent\textbf{Datasets.} We validate the effectiveness of the model on three tasks: hallucination detection, multi-choice question answering, and open-ended text generation. For hallucination detection, we use the HaluEval QA dataset \cite{li2023halueval}, which contains 10K hallucinated samples annotated by human labelers to evaluate the model's ability to recognize and avoid generating hallucinations. For multi-choice question answering, we use the WIKI-FACTOR and NEWS-FACTOR datasets from the FACTOR benchmark \cite{muhlgay2023generating}. WIKI-FACTOR is based on the Wikipedia section of The Pile’s validation split and consists of 2994 examples, while NEWS-FACTOR is based on Reuters articles and consists of 1036 examples. These datasets are designed to test the model's factual reasoning and comprehension abilities. For open-ended text generation, we follow previous work \cite{chuang2023dola} and use the TruthfulQA \cite{lin2021truthfulqa} dataset.

\noindent\textbf{Evaluation Metrics.} For HaluEval, WIKI-FACTOR, and NEWS-FACTOR, we used accuracy as the evaluation metric of hallucination discrimination results. For TruthfulQA, we evaluate the model's performance on multiple choice question answering tasks using MC1, MC2, and MC3. In the open text generation task, since the fine-tuning method used in \cite{lin2021truthfulqa} is no longer available, we used the same fine-tuning data to fine-tune GPT3 to obtain two "GPT judges" to evaluate the accuracy and information content of the output results respectively.

\noindent\textbf{Models.} We used LLAMA2-7B-chat and LLAMA2-13B-chat \cite{touvron2023llama} as the base model to evaluate the consistency of our fine-tuned "GPT-judges" with the original evaluation method. We also introduced the recently released LLAMA3-8B \cite{dubey2024llama} and Qwen2-7B \cite{yang2024qwen2} for comparison with these models to evaluate the ability of existing methods to alleviate hallucinations on the latest large language models.

\noindent\textbf{Baselines.} We compare our model to the following baselines: \underline{(1)} original Decoding (or greedy decoding); \underline{(2)} Dola \cite{chuang2023dola}, which subtracts the final layer logits from the logits of the earlier contrast layer to get the adjusted probability distribution of the next word; \underline{(3)} Activation Decoding (AD) \cite{chen2024context}, that adjusts the probability distribution of the next word according to the sharpness degree of the activation of the next token candidate. Each of these baselines uses only internal representations of the model to help decode and mitigate hallucinations, without the need for external information and extra training.

\subsection{Results and Analysis}

The experimental results presented in Table \ref{tab1} provide a comprehensive evaluation of the proposed model's capabilities in mitigating hallucinations and improving factual accuracy in generated text across different datasets and model scales.

\begin{table}[tbp]
  \centering
  \resizebox{0.6\linewidth}{!}{
    \begin{tabular}{lcc|ll}
    \toprule
    \multirow{2}[4]{*}{\textbf{Model}} & \multicolumn{2}{c}{\textbf{FACTOR}} & \multicolumn{1}{c}{\textbf{HaluEval}} \\
\cmidrule{2-5}          & \textbf{News} & \textbf{Wiki} & \multicolumn{1}{c}{\textbf{QA}}  \\
    \midrule
    LLaMa2-7B-chat + Dola \citep{chuang2023dola} & 48.1  & \textbf{56.5}  &   \multicolumn{1}{c}{51.3}    \\
    + AD \citep{chen2024context} & 61.7  & 53.8  &  \multicolumn{1}{c}{52.4}  \\
    + Ours & \textbf{62.9}  & 55.1  & \multicolumn{1}{c}{\textbf{52.8} }        &  \\
    \midrule
    LLaMa2-13B-chat + Dola \citep{chuang2023dola} & 50.6  & 49.1  & \multicolumn{1}{c}{49.4}     \\
    + AD \citep{chen2024context} & 67.8  & 58.4  &   \multicolumn{1}{c}{49.0} \\
    + Ours & \textbf{68.7}  & \textbf{60.3}  &    \multicolumn{1}{c}{\textbf{49.7} }       &  \\
    \midrule
    LLaMa3-8B + Dola \citep{chuang2023dola} & 60.3  & \textbf{55.7}  &   \multicolumn{1}{c}{35.9}    \\
    + AD \citep{chen2024context} & 59.9  & 48.2  &    \multicolumn{1}{c}{35.7}  \\
    + Ours & \textbf{64.3}  & 50.7  &    \multicolumn{1}{c}{\textbf{36.2} }    \\
    \bottomrule
    \end{tabular}%
  }
  \caption{Performance Comparison on FACTOR and HaluEval Datasets.}
    \label{tab1}%
\end{table}%

When examining the performance on the FACTOR dataset, we observe that our approach almost all outperforms the Dola and AD baselines across all tested model configurations. Notably, the improvement is more pronounced in the News section, where the model's ability to synthesize and fact-check information against a structured narrative is crucial. For instance, our method achieved a notable improvement with the LLaMa3-8B model, increasing the accuracy from 59.9 (AD) to 64.3. This suggests that our method's ability to leverage contextual coherence is particularly effective in domains that require strict adherence to factual information. Similarly, in the Wiki section, we see a positive trend where our method surpasses the baselines, albeit with a smaller margin compared to the News section.

Next, we assessed the model's ability to detect hallucinations using the HaluEval dataset. The HaluEval dataset contains 10,000 hallucinated samples and is designed to evaluate the model's capability to distinguish between factual and non-factual statements \cite{li2023halueval}. On the HaluEval QA dataset, our method again demonstrates its superiority over the baselines, achieving the highest accuracy across all model sizes. This further confirms our method's effectiveness in discerning and avoiding hallucinations, a key aspect of reliable question-answering systems.

The results from the TruthfulQA dataset, as shown in Table \ref{tab:truthfulqa}, provide further insights into the effectiveness of our model in generating factually accurate content while minimizing hallucinations in both multiple-choice (MC) question-answer and open-ended text generation tasks.

\begin{table*}[tbp]
  \centering
  \resizebox{\linewidth}{!}{
    \begin{tabular}{llll|llll}
    \toprule
    \multirow{2}[4]{*}{\textbf{Model}} & \multicolumn{3}{c}{\textbf{TruthfulQA(MC)}} & \multicolumn{4}{c}{\textbf{TruthfulQA(Open-Ended Generation)}} \\
\cmidrule{2-8}          & \textbf{ MC1} & \textbf{MC2} & \textbf{MC3} & \textbf{\% Truth↑ } & \textbf{\% Info↑ } & \textbf{\% T * I↑ } & \textbf{\% Reject↓ } \\
    \midrule
    LLaMa2-7B-chat & 33.6  & 51.3  & 24.9  & 48.4  & \textbf{78.6}  & 30.3  & \textbf{21.5} \\
    + Dola \citep{chuang2023dola} & 29.7  & 51.8  & 21.6  & 48.6  & 78.3  & 30.3  & 21.8 \\
    + AD \citep{chen2024context} & 34    & 51.6  & 25.8  & 47.4  & 80.2  & 31.4  & 18.2 \\
    + Ours & \textbf{34.5}\textcolor{red}{$\uparrow$ 0.9}  & \textbf{52.0}\textcolor{red}{$\uparrow$ 0.7}  & \textbf{26.2}\textcolor{red}{$\uparrow$ 1.3}  & \textbf{48.7}\textcolor{red}{$\uparrow$ 0.3}  & \textbf{78.6}\textcolor{blue}{-}  & \textbf{30.7}\textcolor{red}{$\uparrow$ 0.4} & \textbf{21.5}\textcolor{blue}{-} \\
    \midrule
    LLaMa2-13B-chat & \textbf{35}    & 53.3  & 26.6  & 50.9  & \textbf{84.9}  & 38.5  & \textbf{13.6} \\
    + Dola \citep{chuang2023dola} & 27.1  & 45.8  & 22.9  & 49.6  & 78.1  & 38    & 0.7 \\
    + AD \citep{chen2024context} & 34    & 53.5  & 26.6  & 50.9  & 84.3  & 37.9  & 13.7 \\
    + Ours & 33.9\textcolor{green}{$\downarrow$ 1.1}  & \textbf{53.7}\textcolor{red}{$\uparrow$ 0.4}  & \textbf{27.2}\textcolor{red}{$\uparrow$ 0.6}  & \textbf{51.1}\textcolor{red}{$\uparrow$ 0.2}  & \textbf{84.9}\textcolor{blue}{-}  & \textbf{38.7}\textcolor{red}{$\uparrow$ 0.2}  & \textbf{13.6}\textcolor{blue}{-} \\
    \midrule
    LLaMa3-8B & \textbf{39.9}  & 57.3  & 29.6  & 47.9  & \textbf{56.9}  & \textbf{21.0}    & 32.7 \\
    + Dola \citep{chuang2023dola} & 34.4  & 53.8  & 24.9  & 47.8  & 55.8  & 20.8  & 33.2 \\
    + AD \citep{chen2024context} & 33.9  & 56.9  & 28.9  & 48.7  & 56.1  & 21.3  & 30.5 \\
    + Ours & 39.4\textcolor{green}{$\downarrow$ 0.5}  & \textbf{60.3}\textcolor{red}{$\uparrow$ 3.0}  & \textbf{32.7}\textcolor{red}{$\uparrow$ 3.1}  & \textbf{48.4}\textcolor{red}{$\uparrow$ 0.5}  & 56.5\textcolor{green}{$\downarrow$ 0.4}   &\textbf{21.0} \textcolor{blue}{-}  & \textbf{33.7}\textcolor{red}{$\uparrow$ 1.0}  \\
    \midrule
    Qwen2-7B & \textbf{42.8}  & 61.1  & 32.7  & /     & /     & /     & / \\
    + Ours & 41.5\textcolor{green}{$\downarrow$ 1.3}   & \textbf{61.2}\textcolor{red}{$\uparrow$ 0.1}  & \textbf{33.3}\textcolor{red}{$\uparrow$ 0.6}   & /     & /     & /     & / \\
    \bottomrule
    \end{tabular}%
  }
  \caption{Performance Comparison on TruthfulQA Datasets.}
  \label{tab:truthfulqa}%
\end{table*}%
In the multiple-choice setting, we note modest improvements across all three metrics (MC1, MC2, MC3) when applying our method to the LLaMa2-7B-chat model. The improvements are more pronounced in the MC3 metric, which suggests that our approach is particularly adept at enhancing the model's performance in scenarios requiring a higher degree of factual precision. This trend is similarly observed with the LLaMa2-13B-chat and LLaMa3-8B models, where our method leads to improvements in the MC2 and MC3 metrics, reinforcing the notion that our approach scales effectively with model size.

For open-ended text generation, the "\% Truth" and "\% Info" metrics quantify the truthfulness and informativeness of the generated content, while "\% T * I" represents the combined score, and "\% Reject" indicates the rate at which generated content is deemed unacceptable. Here, our method shows stability in maintaining or slightly improving the truthfulness and combined truthfulness and informativeness scores. However, it is important to note that for the LLaMa3-8B model, there is a slight increase in the "\% Reject" rate, suggesting that while our method aids in factual accuracy, it may also lead to a marginal uptick in generating responses that fall outside acceptable parameters.

The Qwen2-7B model, being the latest addition, provides a baseline for our method's applicability to newer architectures. Even though there is a slight decrease in the MC1 metric, there is a consistent improvement in the MC3 metric, which is critical for evaluating the model's ability to generate nuanced and factually dense responses.

It is worth mentioning that the effectiveness of our approach is not only reflected in the numerical improvements but also in the ability to maintain performance across a range of tasks and model complexities. The results suggest that our method successfully guides the model towards a better understanding of context, which is essential for the nuanced task of generating factually accurate and relevant content. Our method shows promising potential in its application to large language models, emphasizing its capacity to enhance the generation of fact-based responses while curbing the propensity for hallucinations. The experimental outcomes corroborate the viability of our approach as an integral mechanism for improving the quality and reliability of language model outputs, which is essential for their deployment in real-world settings.

\subsection{Qualitative Study}
After comparing our method with the baselines, we verify whether our approach can effectively quantify the amount of information contained in the context tokens and test the effectiveness of our information loss criterion. 

\textbf{Validation of Contextual Information Measurement}

\begin{figure}[htbp]
    \centering
    \begin{subfigure}[b]{0.5\textwidth}
        \centering
        \includegraphics[width=\linewidth]{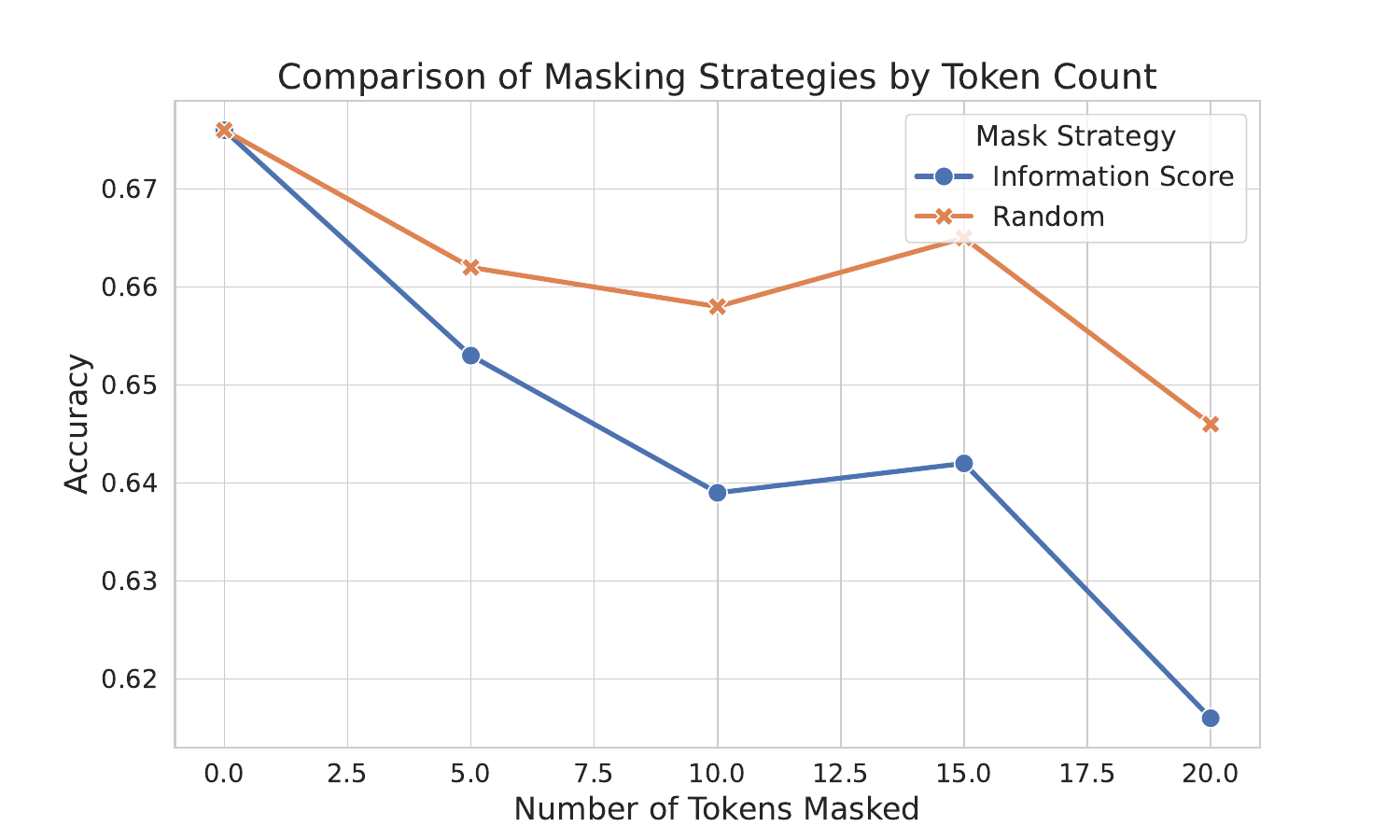}
        \caption{Comparison of Masking Strategies by Token Count. The figure illustrates the decline in model performance when key tokens are masked based on their information content (Information Score) versus randomly masked tokens.}
        \label{fig:mask}
    \end{subfigure}
    \hfill
    \begin{subfigure}[b]{0.45\textwidth}
        \centering
        \includegraphics[width=\linewidth]{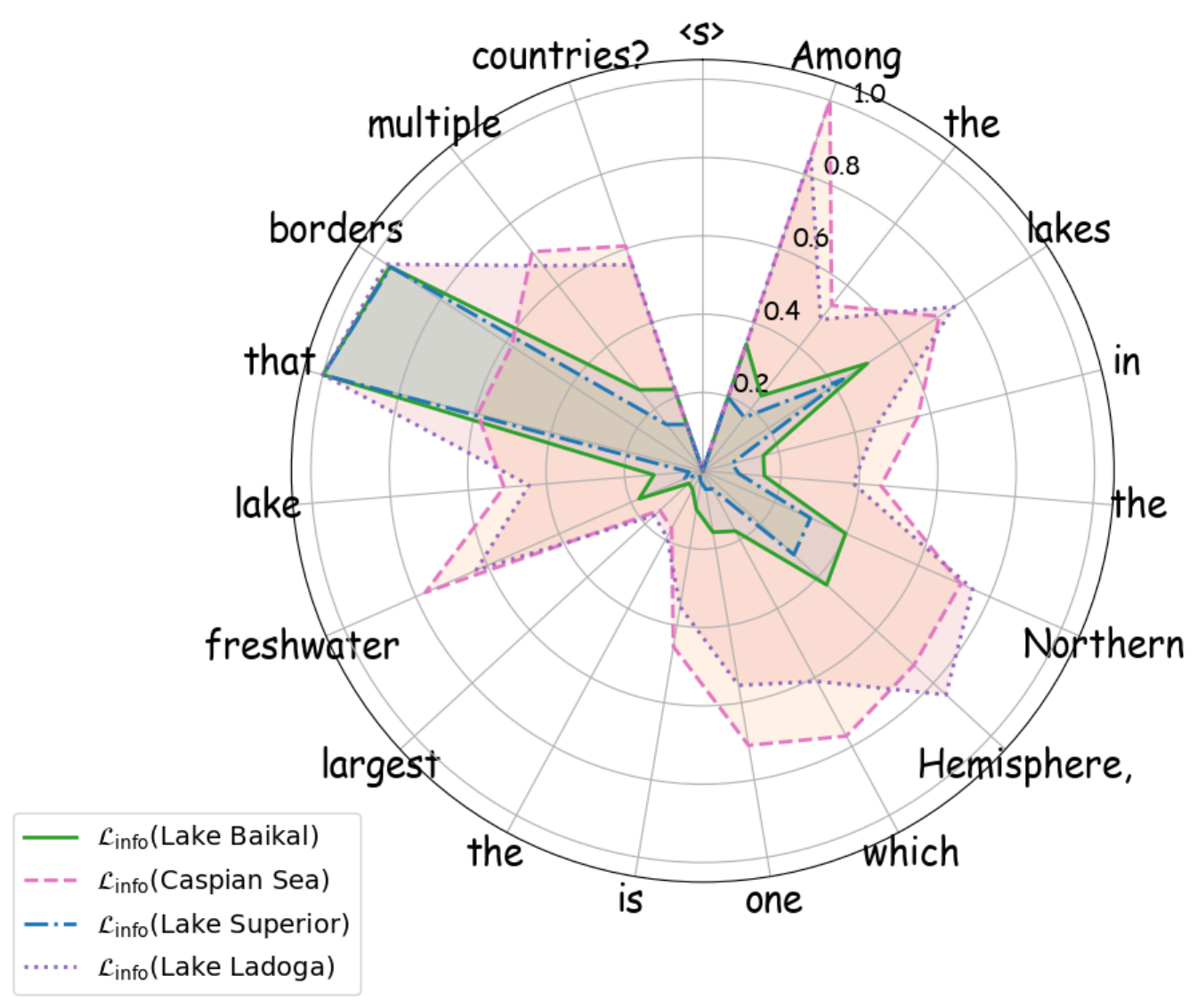}
        \caption{Information Loss Metrics for Different Responses.}
        \label{fig:radar}
    \end{subfigure}
    \caption{Comparison of Masking Strategies and Information Loss Metrics.}
    \label{fig:combined}
\end{figure}

First, we conducted tests on a subset of the FACTOR News dataset to evaluate whether the method proposed in this paper can effectively measure the information content of tokens in the context. Specifically, we randomly selected 500 question-answer samples from the FACTOR News dataset, each containing a question, a correct answer, and three incorrect answers. We then used the AMC-based information measurement method proposed in this paper to assess the information content of all tokens in the questions. Subsequently, we replaced the top 0, 5, 10, 15, and 20 tokens ranked by information content with token "[MASK]", resulting in four questions with varying degrees of information distortion. Finally, we evaluated the importance of the masked words by examining the decline in hallucination discrimination performance of the llama2-7B-chat model on these questions with different degrees of information distortion. In Figure \ref{fig:mask}, we compared these results with those obtained by randomly masking an equal number of tokens.

The results indicate that the masking strategy based on the information score has a more significant impact on model performance compared to the random masking strategy. As the number of key tokens masked increases, the model's performance gradually declines. In contrast, the random masking strategy may only mask tokens with weak semantic associations, such as auxiliary words and prepositions, which have limited impact on the model's performance. Therefore, through the above experiments, we conclude that the AMC-based information measurement method proposed in this paper can indeed effectively characterize the importance of tokens in the context. This helps explain the excellent performance of the hallucination mitigation method based on this approach.

\textbf{Visualization of Information Loss}
Next, we use an example to demonstrate the effectiveness of the information loss metric we have calculated. Consider the question: "Among the lakes in the Northern Hemisphere, which one is the largest freshwater lake that borders multiple countries?" The correct answer is Lake Superior, while incorrect answers include Lake Baikal, Caspian Sea, Lake Ladoga, Qinghai Lake, and Lake Thun. Figure \ref{fig:radar} illustrates the degree of key information loss in different answers. We found that almost all answers are associated with the semantics of "largest" in various contexts, so there is only a small amount of information loss on this key semantic. However, the results in the figure indicate that incorrect answers lead to higher information loss and often result from an overemphasis on a particular piece of key information, causing significant loss in other key areas. For example, the Caspian Sea is the largest saltwater lake, which contradicts the key information "freshwater" in the question, resulting in high information loss in this aspect. In contrast, the correct answer, Lake Superior, aligns well with the description in the question, showing lower information loss across almost all keywords compared to the other incorrect answers. This example demonstrates the rationality and interpretability of our method.

\section{Conclusion}

In this paper, we tried to migrate the phenomenon of hallucinations in Large Language Models (LLMs), which is a critical challenge. Motivated by the need to enhance the reliability and trustworthiness of information generated by LLMs, we proposed a novel decoding strategy that leverages the theory of absorbing Markov chains to mitigate hallucinations. Our approach recognizes the importance of maintaining a coherent and accurate flow of information throughout the generation process, from the initial context to the final output.

By modeling the process of generating new content as an information flow problem, we were able to precisely quantify the significance of contextual information and measure the extent of information loss during the generation process. Our method considers all possible paths from the first to the last token, allowing for a comprehensive assessment of the importance of key information and its propagation. This innovative use of absorbing Markov chains provides a robust mechanism to ensure that the model remains focused on generating factually accurate and contextually coherent content.

Evaluations on diverse datasets, including TruthfulQA, FACTOR, and HaluEval, demonstrated the superior performance of our method in reducing hallucinations and improving the quality of the generated text. These findings underscore the potential of our method to serve as a valuable tool in enhancing the reliability and trustworthiness of LLMs across various applications, from news writing to knowledge dissemination and beyond.

In conclusion, our work highlights the necessity of ensuring accurate information flow in LLMs and offers a promising direction for future research and development in the field of natural language processing. Through continuous innovation and exploration, we aim to further advance the capabilities of LLMs, making them more reliable and trustworthy sources of information for users around the world.


\bibliographystyle{unsrt}  
\bibliography{main}  

\end{document}